\documentclass[conference]{IEEEtran}
\IEEEoverridecommandlockouts

\usepackage{cite}
\usepackage{amsmath,amssymb,amsfonts}
\usepackage{algorithmic}
\usepackage{graphicx}
\usepackage{textcomp}
\usepackage{xcolor}
\usepackage{multirow}
\usepackage{enumitem}
\usepackage{booktabs}
\def\BibTeX{{\rm B\kern-.05em{\sc i\kern-.025em b}\kern-.08em
    T\kern-.1667em\lower.7ex\hbox{E}\kern-.125emX}}
\begin{document}

\title{Less but Better: Parameter-Efficient Fine-Tuning of Large Language Models for Personality Detection
\thanks{This work was supported by the Alan Turing Institute and DSO National Laboratories under a grant on improving multimodal misinformation detection through affective analysis. Additional support was provided by the Interdisciplinary Research Pump-Priming Fund, University of Southampton.}
}

\author{
\IEEEauthorblockN{Lingzhi Shen}
\IEEEauthorblockA{
\textit{University of Southampton} \\
Southampton, United Kingdom \\
l.shen@soton.ac.uk}
\and
\IEEEauthorblockN{Yunfei Long}
\IEEEauthorblockA{
\textit{Queen Mary University of London} \\
London, United Kingdom \\
yl20051@essex.ac.uk}
\and
\IEEEauthorblockN{Xiaohao Cai}
\IEEEauthorblockA{
\textit{University of Southampton} \\
Southampton, United Kingdom \\
x.cai@soton.ac.uk}
\and
\IEEEauthorblockN{Guanming Chen}
\IEEEauthorblockA{
\textit{University of Southampton} \\
Southampton, United Kingdom \\
gc3n21@soton.ac.uk}
\and
\IEEEauthorblockN{Imran Razzak}
\IEEEauthorblockA{
\textit{Mohamed bin Zayed University of Artificial Intelligence} \\
Abu Dhabi, United Arab Emirates \\
imran.razzak@mbzuai.ac.ae}
\and
\IEEEauthorblockN{Shoaib Jameel}
\IEEEauthorblockA{
\textit{University of Southampton} \\
Southampton, United Kingdom \\
M.S.Jameel@southampton.ac.uk}
}

\maketitle

\begin{abstract}
Personality detection automatically identifies an individual's personality from various data sources, such as social media texts. 
However, as the parameter scale of language models continues to grow, the computational cost becomes increasingly difficult to manage. Fine-tuning also grows more complex, making it harder to justify the effort and reliably predict outcomes. We introduce a novel parameter-efficient fine-tuning framework, PersLLM, to address these challenges. In PersLLM, a large language model (LLM) extracts high-dimensional representations from raw data and stores them in a dynamic memory layer. PersLLM then updates the downstream layers with a replaceable output network, enabling flexible adaptation to various personality detection scenarios. By storing the features in the memory layer, we eliminate the need for repeated complex computations by the LLM. Meanwhile, the lightweight output network serves as a proxy for evaluating the overall effectiveness of the framework, improving the predictability of results. Experimental results on key benchmark datasets like Kaggle and Pandora show that PersLLM significantly reduces computational cost while maintaining competitive performance and strong adaptability.
\end{abstract}


\section{Introduction}
Personality refers to the stable traits in an individual’s emotions, thoughts, and behaviours that shape how they perceive, interpret, and interact with the world \cite{cervone2022personality, teng2022understanding}. It is a complex and multifaceted construct that combines various characteristics to form a person's unique identity. The Myers-Briggs Type Indicator (MBTI) \cite{choong2021predicting} is a widely used personality assessment tool based on Carl Jung’s theory of psychological types. It categorizes individuals into 16 distinct personality types by evaluating them across four dimensions: Extraversion (\textbf{E}) vs. Introversion (\textbf{I}), Sensing (\textbf{S}) vs. Intuition (\textbf{N}), Thinking (\textbf{T}) vs. Feeling (\textbf{F}), and Judging (\textbf{J}) vs. Perceiving (\textbf{P}). In recent years, automated personality detection using machine learning has gained widespread attention in fields such as personalized marketing \cite{jenifa2024effective}, mental health assessment \cite{iyortsuun2023review}, and employee recruitment \cite{nirmala2024personality}. 

As consistently demonstrated \cite{zhao2025leveraging, shen2025gamed}, the ability of language models to capture complex linguistic patterns and contextual dependencies is crucial for understanding personality traits from textual data \cite{killian2024detecting}. These models are often extended through continued pre-training \cite{lu2023sentiment} and full fine-tuning \cite{lu2025multimodal} to adapt to specific domains or optimize for particular tasks. LLMs with billions or even trillions of parameters, represent a qualitative leap over traditional models. They have become increasingly popular in tasks involving complex reasoning \cite{minaee2024large} and deep semantic analysis \cite{zhu2023personality}. 
However, as the number of parameters increases, LLMs 
face significant challenges in fine-tuning. These difficulties include enormous computational resource demands, longer training durations, and increased memory consumption, all of which severely impact the ability to comprehensively evaluate an LLM's performance on specific tasks \cite{li2023task}. Additionally, the inability to quickly receive feedback and validate improvements during the development phase further complicates the process. 

Parameter-efficient fine-tuning (PEFT) \cite{hui2024hft} is a practical and effective knowledge transfer strategy that enables large pre-trained models to adapt to new tasks while significantly reducing computational and memory costs. Rather than updating all parameters, PEFT typically optimizes only a small subset. For example, by applying low-rank reparameterization techniques such as LoRA \cite{wu2024lora}, introducing lightweight modules such as adapters \cite{he2021towards}, or incorporating soft prompts such as prefix tuning \cite{li2021prefix} and prompt tuning \cite{jia2022visual}. Recent work like AdapterHub \cite{poth2023adapters} has unified these techniques under a generalized “adapter method” abstraction, further enhancing parameter efficiency and modular composability. This makes it particularly suitable for deployment in resource-constrained environments.
PEFT has been widely adopted in natural language processing, where LLMs can be efficiently customized for specialized applications such as medical diagnosis, legal research, and customer service.

In this work, we develop a novel PEFT framework for personality detection tasks called PersLLM (\textbf{Pers}onality \textbf{L}arge \textbf{L}anguage \textbf{M}odel). Our intuition is that in traditional end-to-end models, each time a new input is provided to the LLM, the model must repeat complex computations \cite{sainath2021efficient}, such as feature extraction, forward propagation, and backpropagation. These steps involve extensive matrix operations and multi-layer computations, such as the self-attention mechanism in transformer architectures and gradient calculations, all requiring access to the vast parameters of the LLM, making it a highly computationally intensive process \cite{dave2021hardware}.
To address this issue, we introduce a novel dynamic memory layer after feature extraction. This layer leverages an indexing mechanism, a querying mechanism, and cache updates to store extracted features, eliminating the need for costly LLM computations during each inference. Besides the feature memory capability, another novelty lies in our choice of a lightweight neural network for the output layer. Compared to the LLM, these networks have fewer parameters and lower computational complexity, significantly reducing the computational cost of training. Unlike training from scratch, the downstream network only needs to learn task-specific relationships and patterns based on the features stored in the memory layer, further accelerating the training process. Additionally, since these output layers are replaceable, the framework gains adaptability. For different personality detection scenarios, we do not need to retrain the LLM. Instead, we can adjust the output network, selecting the appropriate one based on the performance of the individual output layers. 

\noindent \textbf{Main Contributions}: The PersLLM framework not only effectively mitigates the computational overhead and time costs associated with LLMs through PEFT, but also delivers more effective quantitative results. This introduces a novel strategy for fine-tuning LLMs. The key technical contribution is the use of a dynamic memory layer to store and update features, optimizing computational efficiency, while a flexible output network can quickly adapt to various personality detection scenarios. PersLLM has also fully demonstrated the immense potential of fine-tuning LLMs for this specific task. PersLLM outperforms the current state-of-the-art models on two benchmark datasets, setting a new standard for performance.

\section{Related Work}
\noindent \textbf{Classic Personality Detection:} Some commonly used machine learning models such as SVM \cite{Krishna2024AdvancedML} and XGBoost \cite{Salasa2022PersonalityDO} have shown excellent performance in classifying personality traits from social media data. Various deep neural networks also have also demonstrated their strengths in text-based personality detection tasks. A study using BiLSTM with an attention mechanism achieved notable results in detecting psychopathic traits in social media text \cite{Asghar2021DetectionAC}. Graph techniques capture the complex relationships between text and psychological traits by constructing graph structures. For example, both D-DGCN \cite{yang2023orders} and Semi-PerGCN \cite{Zhu2024DataAG} leverage graph convolutional networks to model the structure of user-generated content, with D-DGCN focusing on dynamic relationships between posts, and Semi-PerGCN integrating psycholinguistic features with unsupervised learning. However, most of the above methods often struggle to capture deeper and contextual relationships within unstructured text.

\noindent \textbf{Language Models for Personality Detection:} BERT \cite{alsuhaibani2024idofew}, known for capturing nuanced language patterns, has been fine-tuned for personality recognition \cite{jain2022personality}, such as a study combined sentiment lexicons with the BERT model for multi-label personality detection \cite{Ren2021ASD}. 
Another study used a hybrid strategy to fine-tune RoBERTa for predicting continuous values of personality traits \cite{Wang2024ContinuousOP}. 
Recently, researchers have begun applying LLMs to personality classification tasks. PsyCoT \cite{yang2023psycot} enhances the LLM by using psychological questionnaires as chain-of-thought steps, guiding the model to assess personality traits through a multi-turn dialogue process based on textual input; TAE \cite{hu2024llm} enhances smaller models by extracting knowledge from the LLM through text augmentation, improving post representation and labelling information by combining contrastive learning with semantic, emotional, and linguistic analysis. However, as the size of language models grows, fine-tuning becomes increasingly complex, leading researchers to favour prompt-based methods to reduce training costs. Fine-tuning LLMs for personality detection is even rarer \cite{shen2025ll4g}, which may result in overlooking the most optimal approaches for this task.

Our PersLLM differs from previous approaches as it introduces a PEFT method tailored for personality detection using LLMs. The LLM extracts rich semantic representations, which are stored in a dynamically updated memory layer, while training and prediction are focused on a lightweight and flexible output network. This approach significantly reduces training costs and enhances adaptability to tasks, making LLM fine-tuning feasible for personality detection.

\begin{figure*}[htbp]
    \centering
    \includegraphics[scale=0.160]{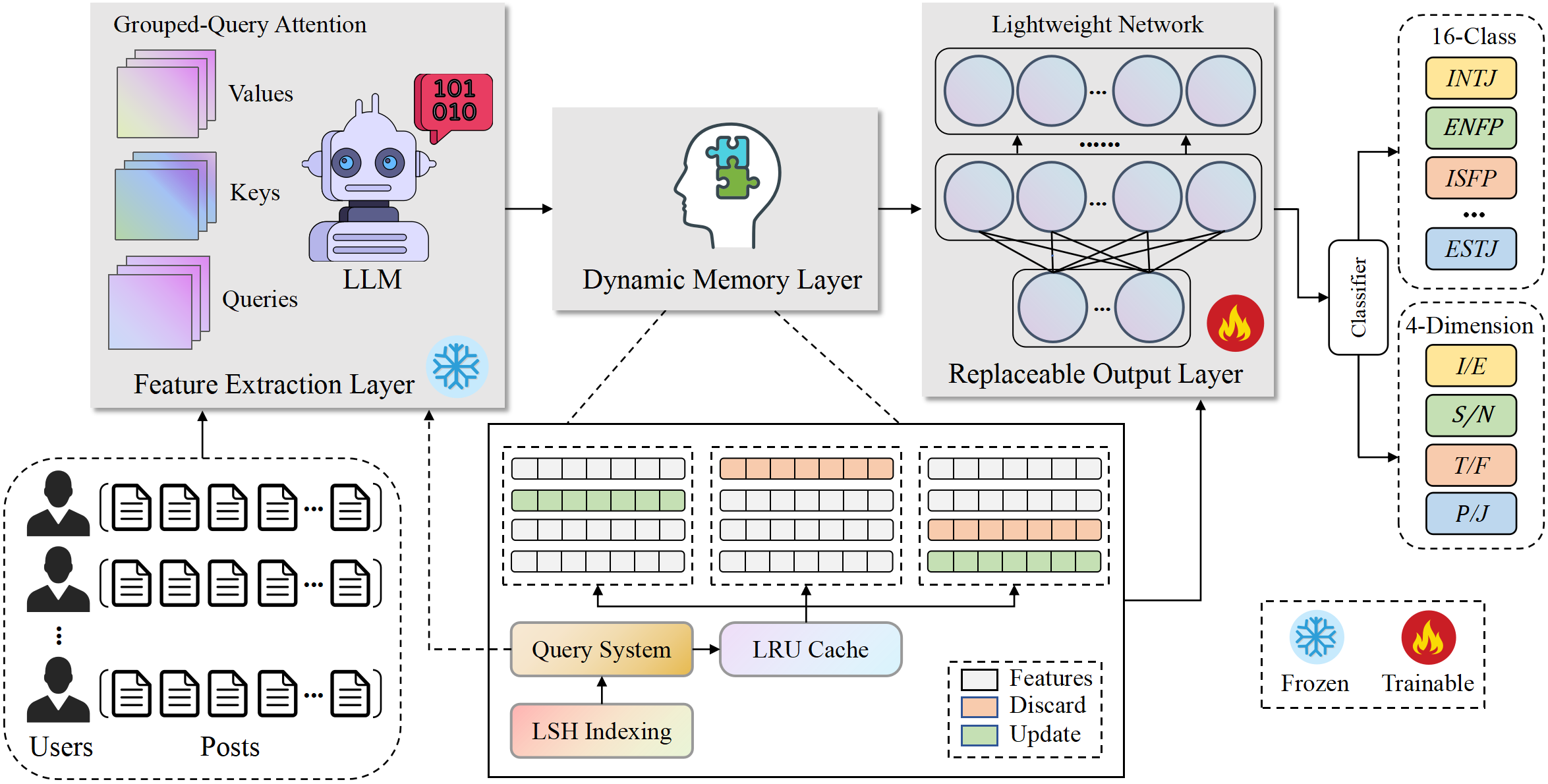} %
    \caption {The framework diagram of PersLLM. First, the LLM is utilized for feature extraction from posts, generating high-dimensional representations that are stored in a dynamic memory layer. Subsequent training is focused entirely on an output network for classification.}
    \label{fig:model-diagram}
\end{figure*}

\section{Methodology}
PersLLM is a novel PEFT framework (in Fig.~\ref{fig:model-diagram}) for personality detection tasks. Leveraging the powerful representation capabilities of the LLM, rich semantic information is extracted from text and stored in a memory layer for repeated access and dynamic updates. The subsequent training and inference processes are handled by a lightweight output network. This design eliminates the computational burden of repeatedly invoking the LLM’s large number of parameters, significantly reducing computational costs while maintaining performance.

\subsection{Feature Extraction Layer}
Under the PersLLM framework, we leverage Llama 3.1 \cite{dubey2024llama} to extract high-dimensional representations from user-generated text. Each user \( u_i \) in the dataset has a collection of multiple posts \( P_i = \{ p_{i1}, p_{i2}, \dots, p_{im} \} \). Since a user may have a large number of posts, they are processed in batches rather than as a single sequence. Each batch \( P_{ib} \subseteq P_i \) consists of a subset of the user’s posts, where multiple batches are processed sequentially.

Llama processes each batch \( P_{ib} \) and generates a corresponding set of feature embeddings \( E_{ib} = \{ e_{ib1}, e_{ib2}, \dots, e_{ibk} \} \), where each \( e_{ibj} \) represents the extracted embedding of the \( j \)-th post in batch \( P_{ib} \). Consequently, instead of a single embedding set per user, we obtain multiple embeddings across batches, i.e., \( E_i = \{ E_{i1}, E_{i2}, \dots, E_{iB} \} \), where \( B \) is the total number of batches for user \( u_i \).

Llama incorporates the Grouped Query Attention (GQA) mechanism to improve efficiency when processing multiple user-generated inputs. GQA reduces the number of independent query heads by a factor of $g$, while keeping the full set of key and value heads. Specifically, the query representation is modified as $Q' \in \mathbb{R}^{s \times (d_k / g)}$, where $Q' = W_Q' \cdot E_{ib}$, $s$ is the total input size, $d_k$ is the original query dimensionality, and $g$ is the number of query groups. The key and value matrices remain unchanged, with $K \in \mathbb{R}^{s \times d_k}$ and $V \in \mathbb{R}^{s \times d_k}$. The attention computation becomes:
\[
\text{Attention}(Q', K, V) = \text{softmax} \left( \frac{Q'K^\top}{\sqrt{d_k / g}} \right) V.
\]

By reducing the dimensionality of queries while maintaining the expressiveness of key-value pairs, GQA lowers the computational overhead without compromising the ability to model long-range dependencies. This allows Llama to efficiently extract contextual representations from multiple posts, facilitating high-performance downstream processing at scale.

After passing through multiple layers of GQA and feedforward neural networks, each batch \( E_{ib} \) is transformed into its high-dimensional feature representation \( H_{ib} = \text{GQA}(E_{ib}) \). Consequently, instead of a single output representation, we obtain a sequence of feature representations across batches, denoted as \( H_i = \{ H_{i1}, H_{i2}, \dots, H_{iB} \} \), where each \( H_{ib} \) encapsulates the extracted features of the corresponding batch.

\subsection{Dynamic Memory Layer}
The dynamic memory layer consists of three core components: an indexing mechanism, a querying mechanism, and a cache update mechanism. Instead of recomputing Llama-extracted representations for every input, the Memory Layer continuously updates its stored features based on recent inputs, leveraging locality-sensitive hashing (LSH) for efficient retrieval.

Specifically, for each extracted feature batch \( H_{ib} \), the Memory Layer applies a LSH transformation to project it into a lower-dimensional space while preserving similarity relationships. LSH approximates nearest neighbor search by mapping similar feature embeddings into the same hash buckets, where the transformation is defined as \( h_{\text{hash}} = \text{sign}(W_{\text{LSH}} H_{ib}) \), with \( W_{\text{LSH}} \) as a random projection matrix that preserves locality properties. The resulting binary vector \( h_{\text{hash}} \) serves as the indexing key for fast retrieval.

Using this hashed representation, the Memory Layer dynamically searches for the closest stored feature batch \( H^* \) by minimizing a predefined distance metric:
\[
H^* = \arg\min_{H_j \in \mathcal{M}} d(H_{ib}, H_j),
\]
where \( \mathcal{M} \) is the memory store and \( d(\cdot, \cdot) \) is a similarity function such as cosine similarity:
\[
S(H_{ib}, H_j) = \frac{H_{ib} \cdot H_j}{||H_{ib}|| \cdot ||H_j||}.
\]

To ensure adaptive retrieval, the Memory Layer dynamically determines whether \( H^* \) can be reused based on a similarity threshold \( \theta \), ensuring that if \( S(H_{ib}, H^*) \geq \theta \), the stored representation is directly used. Otherwise, Llama recomputes the feature embeddings for the batch, and the updated representation is stored for future retrieval.

Given limited storage capacity, the Memory Layer adopts an adaptive Least Recently Used (LRU) strategy, prioritizing frequently accessed embeddings while discarding the least recently used ones when memory is full:
\begin{equation*}
\mathcal{M} =
\begin{cases}
\mathcal{M} \cup \{H_{ib}\}, \ \  \text{if} \  |\mathcal{M}| < M; \\
(\mathcal{M} \setminus H_{\text{oldest}}) \cup \{H_{ib}\}, \ \  \text{if} \ |\mathcal{M}| \geq M.
\end{cases}
\end{equation*}
Here, \( M \) represents the memory capacity, and \( H_{\text{oldest}} \) is the least recently accessed feature batch.

By continuously adapting to new data, the Dynamic Memory Layer optimizes retrieval efficiency while maintaining relevance. Cached feature embeddings evolve dynamically based on recent queries, ensuring that frequently accessed personality representations remain available while outdated ones are discarded.

\subsection{Replaceable Output Layer}
Since the memory layer stores and retrieves embeddings at the batch-level, the output network efficiently transforms the retrieved feature embeddings into compact user-level representations without redundant recomputation.

Given the batch-level feature embeddings \( H_i = \{ H_{i1}, H_{i2}, \dots, H_{iB} \} \) retrieved from the Memory Layer, the output network aggregates them into a final representation suitable for classification. To achieve this efficiently, we adopt a lightweight structure that processes the embeddings in sequence while maintaining adaptability to different architectures.

A computationally efficient sequential model, such as a Gated Recurrent Unit (GRU), serves as an example to illustrate this process. The model iteratively refines an internal representation \( h_t \) at each step \( t \) using the current batch embedding \( H_{it} \) and the prior state \( h_{t-1} \), ensuring that only necessary computations are performed. After processing all batches, the final output representation, denoted as \( h_{\text{final}} \), encapsulates the aggregated information across all stored embeddings.

To maintain efficiency, the Replaceable Output Layer is designed to be lightweight, making it feasible to deploy alternative architectures such as Transformer-based encoders or convolutional networks without significant computational cost. The simplicity of this layer ensures seamless adaptability, enabling it to operate on top of cached feature embeddings without introducing additional complexity.

\subsection{Personality Classification}
The final feature representation, denoted as \( h_{\text{final}} \), obtained from the Replaceable Output Layer, serves as the input to the classification module. Since the output network is replaceable, the classification process remains independent of its specific architecture, ensuring adaptability to different model designs. The classification module performs both binary classification for the four MBTI dimensions and a 16-class classification for predicting the complete personality type.

For the binary classification task, each MBTI dimension \( d \in \{E/I, N/S, T/F, J/P\} \) is treated as an independent classification problem. The model predicts a categorical outcome for each dimension, determining an individual’s preference along the four personality axes, where \( \hat{y} = (\hat{y}_E, \hat{y}_N, \hat{y}_T, \hat{y}_J) \) and each \( \hat{y}_d \) represents the predicted trait for the corresponding dimension.

In addition to binary classification, the model also predicts the complete MBTI personality type as a single categorical variable. This is formulated as a 16-class classification problem, where each class corresponds to one of the 16 MBTI personality types. The model assigns a probability distribution over all 16 personality types and selects the most probable one as the final prediction, where \( \hat{y}_{16} \in \{ 
\text{INTJ}, \allowbreak \text{INTP}, \allowbreak \text{ENTJ}, \allowbreak \text{ENTP}, 
\text{INFJ}, \allowbreak \text{INFP}, \allowbreak \text{ENFJ}, \allowbreak \text{ENFP}, 
\text{ISTJ}, \allowbreak \text{ISFJ}, \allowbreak \text{ESTJ}, \allowbreak \text{ESFJ}, 
\text{ISTP}, \allowbreak \text{ISFP}, \allowbreak \text{ESTP}, \allowbreak \text{ESFP} 
\} \).


\section{Experiments}
\subsection{Experimental Setup}
\noindent \textbf{Datasets:} To ensure a fair comparison with previous work, we selected the same two datasets -- Kaggle and Pandora. The Kaggle dataset is sourced from the PersonalityCafe forum, an online community focused on discussions about personality types. This dataset contains posts from 8,675 users, with each user contributing approximately 45 to 50 posts. The posts cover a variety of topics, including psychology, personal experiences, and everyday discussions. Pandora is a larger corpus from the Reddit platform, which includes MBTI labels for 9,084 users. The number of posts per user varies from dozens to hundreds, and due to the diversity of the Reddit community, the content covers a broader range of topics compared to the Kaggle dataset.

As shown in Table~\ref{tab:data-statistics}, both datasets feature large-scale inputs, user-level aggregation, and significant class imbalance, particularly in the S/N and E/I dimensions. These factors increase the computational burden during training and inference and make it more challenging to capture long-range dependencies and coherent global structures across multiple user posts.

\begin{table}[htbp]
    \caption{Statistics of the Kaggle and Pandora datasets in terms of the set division and class distribution.}
    \scalebox{1.10}
    {
    \centering
    \begin{tabular}{c|c|c|c|c}
        \hline
        \textbf{Dataset} & \textbf{Types} & \textbf{Train} & \textbf{Validation} & \textbf{Test} \\
        \hline
        \multirow{5}{*}{\textbf{Kaggle}} & E/I & 1194 / 4011 & 409 / 1326 & 396 / 1339 \\
        & S/N & 610 / 4478 & 222 / 1513 & 248 / 1487 \\
        & T/F & 2410 / 2795 & 791 / 944 & 780 / 955 \\
        & J/P & 2109 / 3096 & 672 / 1063 & 653 / 1082 \\
        & Posts & 246794 & 82642 & 82152 \\
        \hline
        \multirow{5}{*}{\textbf{Pandora}} & E/I & 1162 / 4278 & 386 / 1427 & 377 / 1437 \\
        & S/N & 727 / 4830 & 208 / 1605 & 210 / 1604 \\
        & T/F & 3549 / 1891 & 1120 / 693 & 1182 / 632 \\
        & J/P & 2229 / 3211 & 770 / 1043 & 758 / 1056 \\
        & Posts & 523534 & 173005 & 174080 \\
        \hline
    \end{tabular}
    }
    \label{tab:data-statistics}
\end{table}

\begin{table*}[htbp]
    \caption{Comparison of PersLLM with state-of-the-art language model-based baselines on the Kaggle and Pandora datasets in terms of the Macro-F1 (\%) scores across the four dimensions and their overall average (Avg).}
    \centering
    \scalebox{1.00}
    {
    \begin{tabular}{l|cccc|c|cccc|c}
        \hline
        \multirow{2}{*}{\textbf{Methods}} & \multicolumn{5}{c|}{\textbf{Kaggle}} & \multicolumn{5}{c}{\textbf{Pandora}} \\
        \cline{2-11}
        & E/I & S/N & T/F & J/P & \textbf{Avg} & E/I & S/N & T/F & J/P & \textbf{Avg} \\
        \hline
        BERT & 64.65 & 57.12 & 77.95 & 65.25 & 66.24 & 54.22 & 48.71 & 64.70 & 56.07 & 56.56 \\
        RoBERTa & 61.89 & 57.59 & 78.69 & 70.07 & 67.06 & 54.80 & 55.12 & 63.78 & 55.94 & 57.41 \\
        Transformer-MD & 66.08 & 69.10 & 79.19 & 67.50 & 70.47 & 55.26 & 58.77 & 69.26 & 60.90 & 61.05 \\
        PQ-Net & 68.94 & 67.65 & 79.12 & 69.57 & 71.32 & 57.07 & 55.26 & 65.64 & 58.74 & 59.18 \\
        GPT-4 Turbo (Zero-shot) & 68.86 & 54.69 & 80.10 & 66.93 & 67.65 & 57.38 & 51.47 & 71.75 & 62.29 & 60.72 \\
        GPT-4 Turbo (CoT) & 66.63 & 61.85 & 77.23 & 60.80 & 66.62 & 60.97 & 57.14 & 66.39 & 56.92 & 60.36 \\
        DeepSeek-V3 (Zero-shot) & 69.76 & 58.61 & 75.71 & 64.37 & 67.11 & 61.39 & 54.31 & 68.05 & 58.17 & 60.48 \\
        DeepSeek-V3 (CoT) & 68.59 & 61.47 & 74.56 & 61.21 & 66.46 & 58.23 & 56.15 & 67.94 & 55.01 & 59.33 \\
        PsyCoT & 66.56 & 61.70 & 74.80 & 57.83 & 65.22 & 60.91 & 57.12 & 66.45 & 53.34 & 59.45 \\
        TAE & 70.90 & 66.21 & 81.17 & 70.20 & 72.07 & 62.57 & 61.01 & 69.28 & 59.34 & 63.05 \\
        \hline
        \textbf{PersLLM} & \textbf{76.71} & \textbf{75.55} & \textbf{85.11} & \textbf{75.96} & \textbf{78.33} & \textbf{68.89} & \textbf{66.72} & \textbf{73.70} & \textbf{68.55} & \textbf{69.47} \\
        \hline
    \end{tabular}
    }
    \label{tab:macro-f1}
\end{table*}

\noindent \textbf{Implementation Details:} We leveraged the ``Llama-3.1-8B-Instruct'' model. For the output network, we primarily utilized GRU and flexibly replaced it with other neural networks based on task requirements. The feature vector length extracted by Llama and the input vector length of the output network were both standardized to 4096. To maintain lightweight computation, the GRU consists of only three hidden layers by default, and dropout regularization with a probability of 0.2 was applied to prevent overfitting. The hidden dimension of GRU output at each time step is 512. The model uses the Adam optimizer with a learning rate of $1\times10^{-3}$. The loss function used is cross-entropy loss. To prevent information leakage, words that directly matched personality labels were removed during data preprocessing. The datasets were split into training, validation, and test sets using a 60\%/20\%/20\% ratio. The results reported are averages of ten runs.

\noindent \textbf{Evaluation Metrics:} Following previous work, we used Macro-F1 score as the primary evaluation metric, along with accuracy (ACC), precision (P), and recall (R) in the newly added scenarios for comprehensive evaluation.

\noindent \textbf{Baselines:} We selected several widely-used pretrained language models for text classification, including 
fine-tuned BERT \cite{keh2019myers} and RoBERTa \cite{liu2019roberta}, as well as advanced LLMs like GPT-4 \cite{achiam2023gpt} and DeepSeek-V3 \cite{liu2024deepseek}. We also included more complex architectural designs based on different language models. 
The Transformer-MD \cite{yang2021multi} was chosen for its memory-augmented Transformer-XL architecture and dimension-specific attention, enabling effective integration of multi-post inputs for personality detection. The PQ-Net \cite{yang2021learning} model was selected for its dual-stream design that combines text and psychological questionnaires via cross-attention to extract key personality cues.
The PsyCoT \cite{yang2023psycot} was included for its chain-of-thought prompting based on questionnaire items, guiding the LLM through multi-turn reasoning to score each trait. The TAE \cite{hu2024llm} was chosen for leveraging LLM-generated text augmentations and label explanations with contrastive learning to capture psycholinguistic features. 

\subsection{Overall Results}
\noindent \textbf{Macro F1:} The results comparing PersLLM with baselines in Macro-F1 scores are shown in Table~\ref{tab:macro-f1}. PersLLM achieves the top performance across all four dimensions and the overall average. On the Kaggle dataset, the average score is 78.33\%, surpassing the best existing scheme, TAE, by 6.26\%. On the Pandora dataset, PersLLM achieves an average score of 69.47\%, outperforming TAE by 6.42\%. PersLLM also delivers outstanding results across all four dimensions. Compared to the baselines, PersLLM shows significant improvements in dimensions with severe class imbalance, particularly narrowing the performance gap between T/F and the other three dimensions.

Compared to previously reported Transformer-based language models such as BERT and RoBERTa, the improvements observed in PersLLM indicate that increasing the number of parameters significantly enhances the model's ability to capture richer semantic representations and handle deeper linguistic nuances. We also compared PersLLM with several contemporary LLM-based approaches and found that PersLLM consistently outperforms methods that rely on prompting, such as zero-shot. Surprisingly, chain-of-thought (CoT) reasoning with step-by-step inference produced results that were even slightly worse than zero-shot performance. This finding suggests that the model's capacity constraints limit its ability to further push performance boundaries, and the complex decomposition process in CoT can even introduce hallucinations and noise. When the model's discriminative power is sufficiently strong, adjusting its parameters should be prioritized to break through its performance ceiling.

\begin{table}[htbp]
    \caption{The performance comparison of the models within the 16-class evaluation framework.}
    \centering
    \scalebox{1.00}
    {
    \begin{tabular}{cccccc}
        \toprule
        Dataset & Method & Acc & P & R & F1 \\ 
        \midrule
        \multirow{10}{*}{Kaggle}
        & BERT & 34.64 & 16.29 & 14.43 & 15.87 \\
        & RoBERTa & 38.56 & 25.26 & 21.25 & 22.72 \\
        & Llama 3.1 & 43.16 & 37.53 & 32.37 & 34.79 \\
        & Llama 3.1 (LoRA) & 47.24 & 44.15 & 36.62 & 39.43 \\
        & \textbf{PersLLM} & 46.54 & 45.54 & 39.60 & 41.61 \\ 
        & \textbf{PersLLM (GCN)} & 51.14 & 48.60 & \textbf{46.57} & 44.21 \\
        & \textbf{PersLLM (LSTM)} & \textbf{58.43} & \textbf{50.86} & 45.09 & \textbf{47.15} \\
        \midrule
        \multirow{10}{*}{Pandora} 
        & BERT & 27.40 & 5.03 & 8.49 & 5.65 \\
        & RoBERTa & 30.49 & 10.65 & 9.63 & 9.98 \\
        & Llama 3.1 & 37.88 & 45.30 & 24.99 & 30.40 \\
        & Llama 3.1 (LoRA) & 39.51 & 46.38 & 28.71 & 32.88 \\
        & \textbf{PersLLM} & 42.67 & 60.62 & 31.18 & 37.25 \\
        & \textbf{PersLLM (GCN)} & 43.72 & 55.52 & \textbf{31.73} & 36.86 \\
        & \textbf{PersLLM (LSTM)} & \textbf{45.59} & \textbf{69.67} & 30.49 & \textbf{38.75} \\
        \bottomrule
    \end{tabular}
    }
    \label{tab:16class}
\end{table}

\begin{figure*}[htbp]
    \centering
    \includegraphics[scale=0.094]{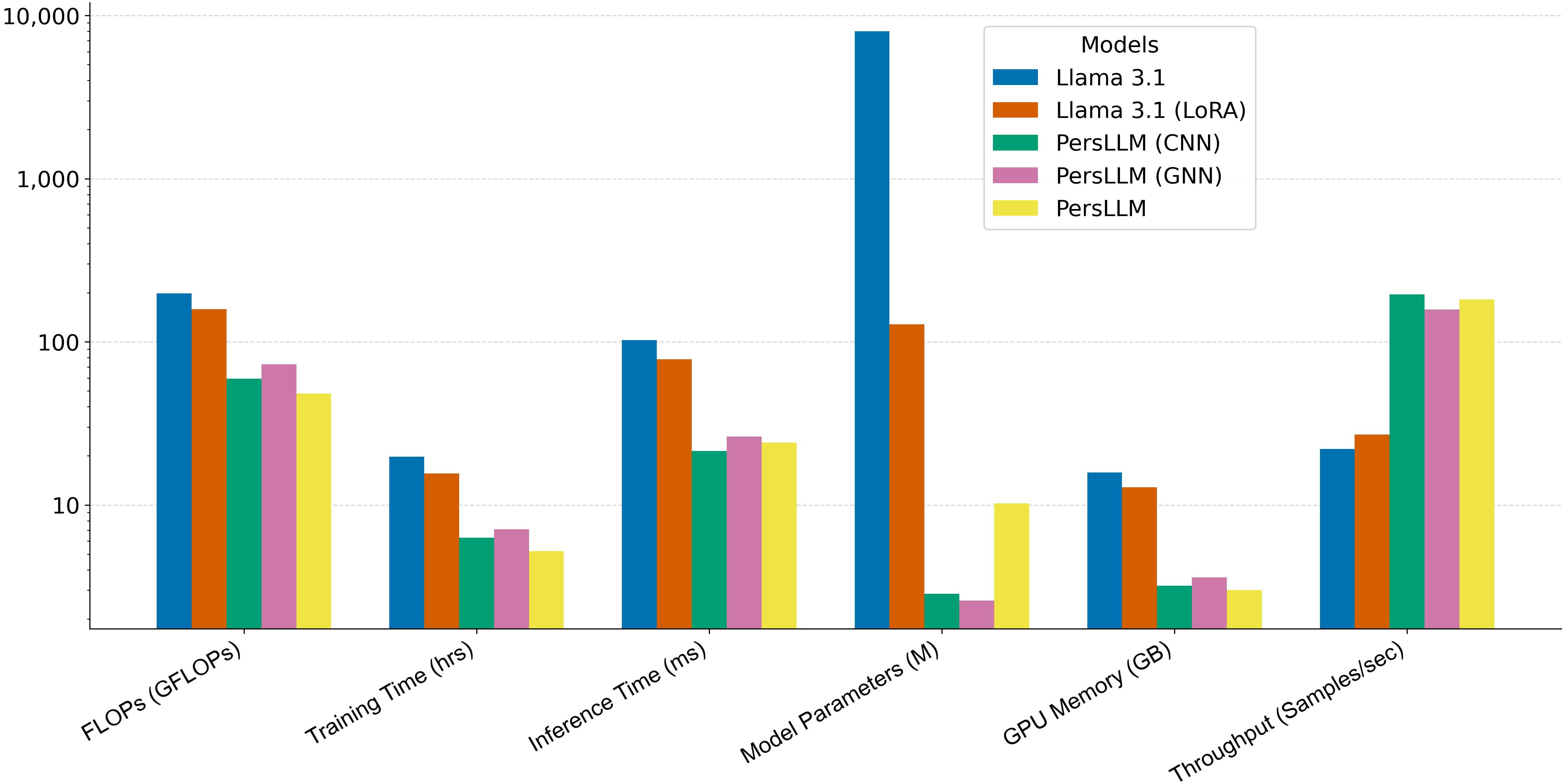} %
    \caption {The paired bar charts in a logarithmic scale for the computational resource usage comparison between our PersLLM, Llama 3.1 and its LoRA fine-tuning on the Kaggle dataset.}
    \label{fig:comparison}
\end{figure*}

\noindent \textbf{16-class Evaluation:} We also developed a new 16-class evaluation framework designed to classify the 16 personality types of the MBTI. Unlike trait-based theories, such as the Big Five personality model \cite{de2000big} that describes personality through continuous trait dimensions, MBTI is grounded in a discrete typology theory. Due to MBTI's emphasis on type distinctions in personality assessment, a specialized classification approach is required to address the challenges in practical applications and optimize costs. 

As shown in Table~\ref{tab:16class}, we selected BERT and RoBERTa as our baselines, as they are widely used language models known for their strong performance on various classification tasks. However, both models performed poorly in the 16-class classification scenario, even achieving near single-digit scores on some metrics for the Pandora dataset. This is primarily due to their lack of task-specific tuning for the 16-class scenario. However, such adjustments often come at a high cost, especially for language models that require full-parameter updates or rely heavily on large-scale labeled data. We compared PersLLM with Llama 3.1 fine-tuned using Low-Rank Adaptation (LoRA) \cite{hu2021lora}. Although LoRA is also an efficient fine-tuning method, its performance is significantly worse across all metrics compared to the various configurations of PersLLM. In contrast, PersLLM can achieve better results by swapping the output network according to the scenario's requirements. For instance, using LSTM on large and complex datasets helps learn more detailed features; or in more sensitive applications, such as medical diagnostics \cite{ochoa2022graph}, using GCN can improve the detection of positive samples. Either replacing LSTM or GCN yields better results than the initial PersLLM, making it more suited for the 16-class classification task. This highlights the flexibility of the PersLLM framework.

\begin{figure}[htbp]
    \centering
    \includegraphics[scale=0.086]{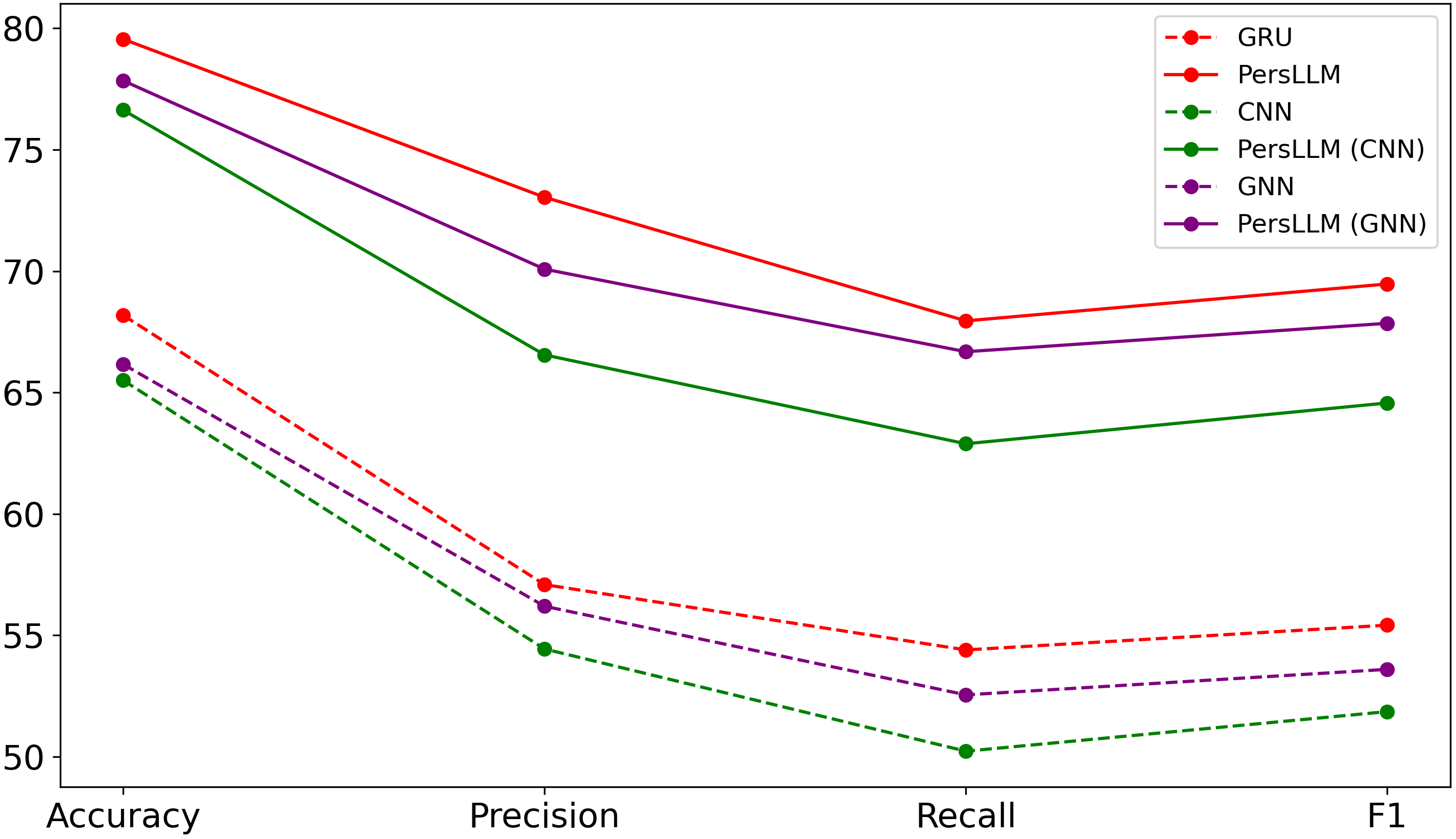}%
    \caption{The line chart reflects the correlation between the performance of the output neural network and the overall performance of PersLLM. The test was conducted on the Pandora dataset.}
    \label{fig:line-chart}
    \vspace{-0.14in}
\end{figure}

\begin{figure*}[htbp]
    \centering
    \includegraphics[scale=0.12]{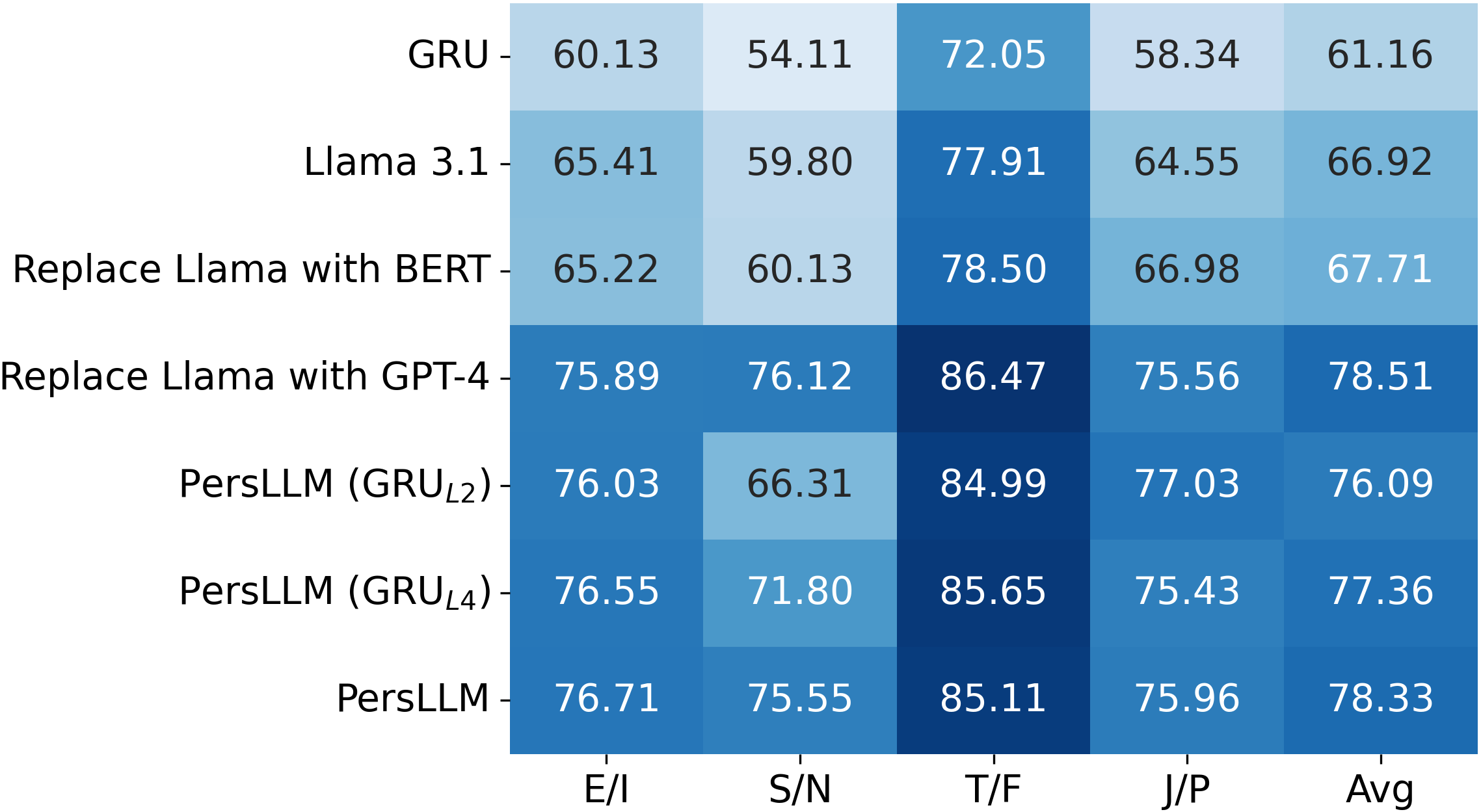}%
    \includegraphics[scale=0.12]{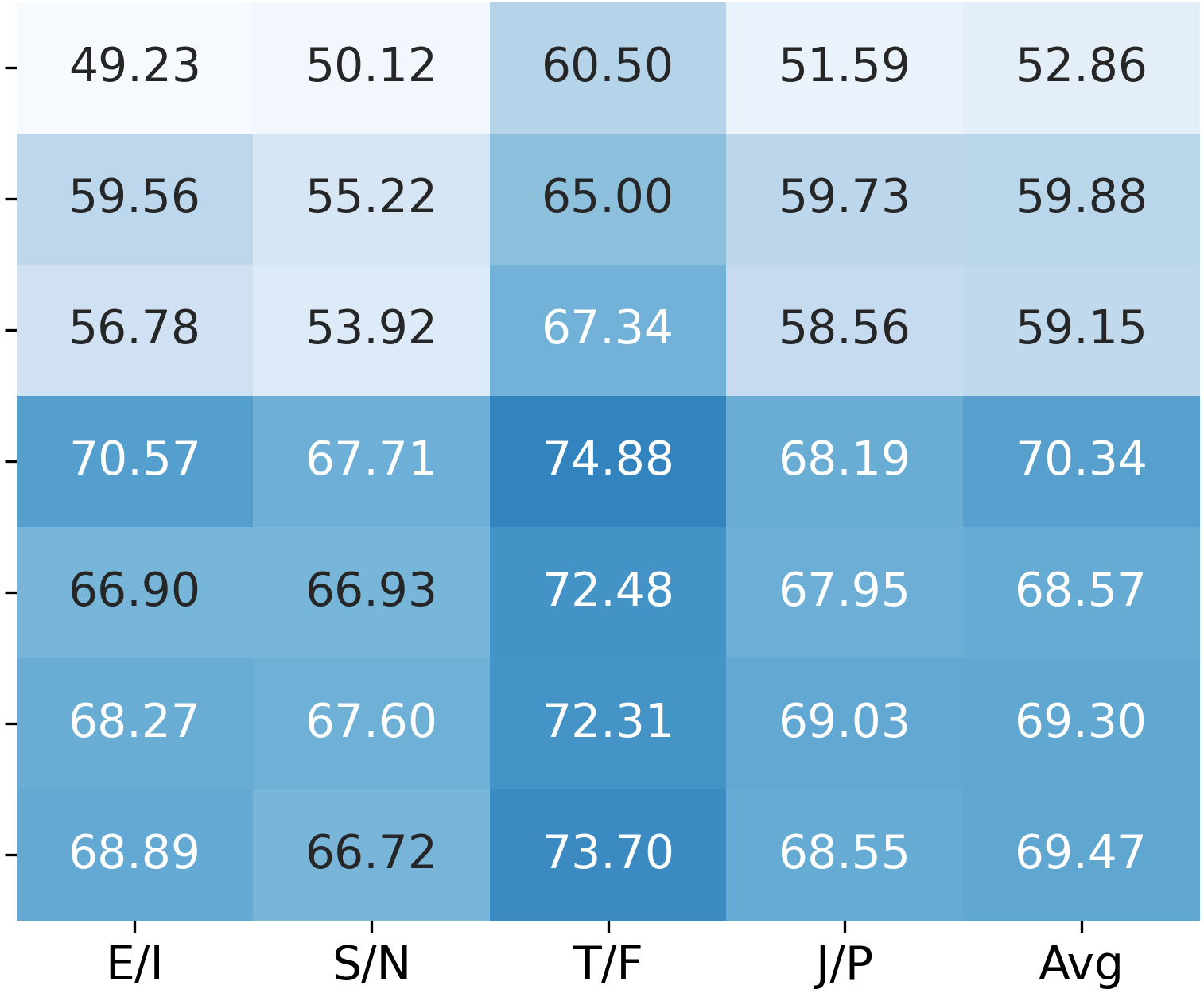}
    \caption{The heatmap for ablation study. It illustrates the performance of various key components of PersLLM on the Kaggle (left) and Pandora (right) datasets. Performance improves as the colour deepens.}
    \label{fig:heatmap}
\end{figure*}

\noindent \textbf{Computational Costs Comparison:} As shown in Fig.~\ref{fig:comparison}, we use paired bar charts to compare the performance of Llama 3.1, LoRA fine-tuning, and our PersLLM, incorporating different output networks such as CNN and GCN, across multiple computational dimensions on the Kaggle dataset, including FLOPs, training time, inference time, model parameters, GPU memory usage, and throughput. Due to the large scale of model parameters, we used a logarithmic scale for resource values. PersLLM demonstrates significant advantages in computational efficiency compared to Llama 3.1 and LoRA fine-tuning, with a notable reduction in FLOPs and GPU memory usage. Additionally, PersLLM requires less training time while maintaining competitive inference times. Interestingly, despite its more compact architecture, PersLLM achieves higher throughput (samples per second), highlighting its excellent ability to balance computational efficiency and performance, effectively addressing the challenges of fine-tuning large models.

As previously mentioned, a key advantage of the PersLLM framework is its ability to assess overall model performance through the output network, thereby improving efficiency and reducing computational costs. As shown in Fig.~\ref{fig:line-chart}, the independent performance of output layers such as GRU, RNN, and CNN on the Pandora dataset is almost identical to their performance when integrated into the PersLLM framework as output networks. This consistency indicates that output networks can reliably serve as proxies to predict PersLLM's final results, allowing us to make stable performance estimates without running the full model. Therefore, in optimization scenarios, we can first experiment with the output network to predict PersLLM's outcomes, ensuring that we can accurately identify necessary adjustments before performing expensive and time-consuming full-scale model tuning.

\subsection{Ablation Study}

In the ablation study, we observed changes in the overall performance of PersLLM by replacing its key components. The results of the ablation analysis are visualized using a heatmap for a more intuitive representation, as shown in Fig.~\ref{fig:heatmap}. 

The heatmap shows that both PersLLM and its variants with replaced components exhibit a significant performance drop on the Pandora dataset compared to the Kaggle dataset (indicated by lighter colours). Similarly, the performance of output modules with different layer configurations declines across all dimensions. This drop may be attributed to inherent differences between the two datasets, such as greater variability or more complex textual patterns in the Pandora dataset. The heatmap highlights the challenge of extending models optimized for one dataset to another without sacrificing performance. Due to the flexibility of PersLLM's PEFT, as previously mentioned, it can quickly adapt to the Pandora dataset by replacing the output layer with other neural networks, such as GRU, without needing to retrain the entire model.

Next, we removed the output layer and used fine-tuned Llama 3.1 with a simple linear layer for direct classification. The average F1 score drops by 11.41\% on Kaggle and 9.59\% on Pandora. The standalone GRU exhibits a more significant decline. Among the four specific dimensions, the largest decline is observed in the S/N dimension. This demonstrates that PEFT significantly enhances the model's adaptability and its ability to capture task-specific nuances. 

To further investigate the impact of different backbone models, we replaced Llama with BERT and GPT-4. BERT performed significantly worse than Llama on both datasets. In contrast, GPT-4 outperformed Llama in most dimensions, demonstrating its strong generalization capability on MBTI personality data.

We also experimented with the output network of varying depths. Our observations show that stacking up to three hidden layers in the GRU maximizes the overall performance of PersLLM. However, adding a fourth layer leads to overfitting, causing a performance drop, highlighting the trade-off between model depth and generalization. Interestingly, the four-layer and three-layer models perform better on different dimensions, suggesting that neural networks of varying depths enable the framework to learn more subtle or specific features on certain dimensions. This points to a clear direction for optimizing model performance: {\it rather than solely increasing depth, selectively adjusting the architecture for specific dimensions can better meet particular needs}. This ability to rapidly tune parameters is a key advantage of our PersLLM architecture. 
This flexibility, combined with targeted fine-tuning, provides a more efficient approach to improving model performance without incurring the computational cost of retraining the entire model.

\section{Conclusions}
We proposed PersLLM, a novel PEFT framework that leverages LLMs for personality detection tasks. The LLM extracts rich representations, which are stored in a dynamic memory layer, avoiding the need for repeated feature computation. The subsequent training and prediction processes focus on a low-parameter output network, which can be replaced based on task requirements. This approach allows for flexible and lightweight fine-tuning of the LLM, significantly reducing computational costs. Experimental results demonstrate that PersLLM outperforms existing state-of-the-art methods by a large margin, proving its effectiveness and efficiency. Future work will incorporate multimodal data, such as images and audio, to enhance the framework's scalability. Additionally, we will develop adaptive fine-tuning techniques that dynamically adjust the training effort based on task requirements, further improving the framework's efficiency.

\vspace{0.10in}


\end{document}